\documentclass[letterpaper]{article} 
\usepackage{aaai25}  
\usepackage{times}  
\usepackage{helvet}  
\usepackage{courier}  
\usepackage[hyphens]{url}  
\usepackage{graphicx} 
\usepackage{amsmath}
\usepackage{natbib}  
\usepackage{caption} 
\usepackage{algorithm}
\usepackage{algorithmic}

\usepackage{newfloat}
\usepackage{listings}
\DeclareCaptionStyle{ruled}{labelfont=normalfont,labelsep=colon,strut=off} 
\floatstyle{ruled}
\newfloat{listing}{tb}{lst}{}
\floatname{listing}{Listing}
\title{From Frustration to Fun: An Adaptive Problem-Solving Puzzle Game Powered by Genetic Algorithm}
\author {
    Matthew McConnell,
    Richard Zhao
}
\affiliations {
    Department of Computer Science, University of Calgary\\
    Calgary, Alberta, Canada, T2N 1N4\\
    matthew.mcconnell1@ucalgary.ca, richard.zhao1@ucalgary.ca
}

\begin{document}

\maketitle

\begin{abstract}
This paper explores adaptive problem solving with a game designed to support the development of problem-solving skills. Using an adaptive, AI-powered puzzle game, our adaptive problem-solving system dynamically generates pathfinding-based puzzles using a genetic algorithm, tailoring the difficulty of each puzzle to individual players in an online real-time approach. A player-modeling system records user interactions and informs the generation of puzzles to approximate a target difficulty level based on various metrics of the player. By combining procedural content generation with online adaptive difficulty adjustment, the system aims to maintain engagement, mitigate frustration, and maintain an optimal level of challenge. A pilot user study investigates the effectiveness of this approach, comparing different types of adaptive difficulty systems and interpreting players' responses.  This work lays the foundation for further research into emotionally informed player models, advanced AI techniques for adaptivity, and broader applications beyond gaming in educational settings. 
\end{abstract}

%

\section{Introduction}
The increased prevalence of online learning intensified issues in students, such as anxiety, stress, and a lack of confidence in learners \cite{adnan2020online}. These online learning based issues have become increasingly prominent, as the absence of face-to-face interaction and personalized support can magnify feelings of isolation and frustration. Traditional online learning environments often adopt a one-size-fits-all approach, failing to account for the diverse emotional and cognitive needs of students \cite{sit2005experiences}.

Recently, research on improving various e-learning elements have gained immense traction, due to the increasing access to technologically-based tools at a large scale. This holds particularly true considering the rapid advancement of artificial intelligence (AI) technologies and various integrations for the Internet of Things (IoT), challenging long-standing traditional models of education. Many of these technological advancements were bred from the COVID-19 pandemic, in which education was forcibly shifted from in-person, classroom-based instruction to online and virtual spaces. As such, a growing need for scalable, robust, and user-centric educational tools has been recognized. 

Virtual learning environments are often identified by explicitly defined information spaces, integrating heterogeneous technologies and pedagogical approaches with the overarching goal of aiding some form of education \cite{VLE}. These defined information spaces can be presented in many forms, including websites, standalone interactive software, or even serious games \cite{garcia2019serious}. Virtual learning environments can often be extended to incorporate AI technologies, such as procedural content generation (PCG) \cite{shaker2016procedural}. PCG techniques can be used to create digital content in real time, which when paired with some form of player or user modeling, can produce adaptive and dynamic systems that adapt to users' needs.

A common method for creating procedurally generated content is genetic algorithms (GA), in which a process mimicking biological evolution is used to find and optimize solutions to various problems \cite{mitsis2020procedural}. These algorithms are based on natural selection, in which a population of individual candidate solutions is modified over many iterations, eventually ``evolving" towards an optimal solution \cite{scirea2020adaptive}.

In this research, we present an Adaptive Problem-Solving Game (APSG), powered by GA and player modeling, in which puzzles are dynamically generated to be tailored to each individual student's difficulty level. The novelty lies in its integration of a genetic algorithm with ''real-time'' puzzle generation, tailored to individual skill levels, unlike traditional adaptive systems which rely solely on offline analysis. Our research addresses three core research questions:

\begin{itemize}
    \item \textbf{RQ1}: Does a real-time genetic algorithm-based puzzle-generation system, informed by player modeling metrics, reduce player frustration more effectively compared to similar approaches?
    \item \textbf{RQ2}: Do puzzles dynamically generated by a customized genetic algorithm informed by real-time player interactions align closely with players' perceived optimal difficulty and provide a clear sense of skill progression?
    \item \textbf{RQ3}: When dynamically adjusting puzzle difficulty in real-time, is ``time-on-task`` alone sufficient to accurately inform the adaptive puzzle-generation process, compared to using multiple player metrics?
\end{itemize}

The paper's primary contribution is a comparative evaluation of adaptive-difficulty approaches, accompanied by an analysis of how players respond to each one. Its secondary contribution is the design of an adaptive puzzle generator that produces tasks across a calibrated difficulty spectrum by means of a customized genetic algorithm.

\section{Related Works}
It is well known that personalized learning materials, particularly in the form of one-on-one tutoring is extremely beneficial to educational success \cite{bloom19842}. Under the best learning conditions that can be devised (personalized tutoring), the average student is two standard deviations above the average control student taught under conventional methods \cite{bloom19842}. However, the cost of personalized education can often be costly, time consuming, and emotionally draining \cite{sharif2022proposed}. Higher levels of technological adaptation in the education sector can have significant impact as it can mitigate common issues with large scale education such as increasing numbers of students, limited public funding, and increased demand for higher-quality education \cite{sharif2022proposed}.

This holds particularly true for online-based education, which has often been forced upon students as part of the new post-pandemic norm. A study by W.H. Sit et al. explored students' views of online learning initiatives, evaluating both the positive and negative experiences of students. They found that while most students were generally on board with online-based education, as it provides time-saving and easy access to material, they desired more personalized learning materials and wished for a more interactive system \cite{sit2005experiences}.

AI has been the topic of many recent research studies, particularly in its relation to education, serious games, and online learning. There exist a multitude of varying techniques, which when used effectively, can improve educational standards \cite{holmes2022state}. Algorithmic approaches, such as rule-based AI \cite{swiechowski2018grail} or biologically-inspired genetic algorithms \cite{darejeh2024framework, scirea2020adaptive} can provide heuristic-based approaches to both content delivery, and adaptivity. Machine learning (ML) based approaches \cite{ciolacu2021education, sharif2022proposed} tend to focus on data-driven models, in which large amounts of data are used to model information surrounding students or players, or analyze various feedback mechanisms in the learning pipeline. Reinforcement learning (RL) approaches \cite{flores2019proposal, kardan2010smart} provide AI systems which can be trained to dynamically adjust learning pathways or game mechanics based on reward-based, trial-and-error interactions. Further, there exists Hybrid models which can either use multiple AI-approaches in tandem, or, pick and choose various individual system components tailored towards specific needs, leveraging the strengths of each, such as rule-based systems for initial scaffolding and ML models for fine-tuned personalization \cite{hare2023combining}.

Garavaglia et al. (\citeyear{garavaglia2022moody5}) proposed personality-biased agents, powered by algorithmically based AI frameworks, that can dynamically adapt their content based on the user's current state. They showcase forms of player modeling, in which emotional states of the user can be analytically read and consequently examined, to update the AI system accordingly. AI techniques have further been used for learning analytics for serious games. Perez-Colado J. et. al. (\citeyear{perez2018multi}) proposed a learning analytics system from the perspective of data-driven user modeling, paying specific attention to educational serious games. They recommend an integrated, user-centric approach, in which educational and game communities must work together to provide complex, multi-level, or hierarchical metrics for analysis. 

Virtual learning environments are digital virtual spaces that facilitate the delivery of curriculum content, assessment, and evaluation activities for students \cite{caprara2022effects}. García-Redondo et al. (\citeyear{garcia2019serious}) explored the impact of a serious game based on multiple intelligences primarily focussed on attention and ADHD, revealing significant improvements in visual attention. A serious game was also proposed to reduce perioperative anxiety and pain in children undergoing ambulatory surgery \cite{verschueren2019development}.

El Khayat et al. (\citeyear{el2012intelligent}) developed an intelligent serious game for children with learning disabilities, focusing on intervention as early as kindergarten, to enhance learner capabilities. They presented an intelligent web-based adaptive serious game, providing us with a strong methodology for tailoring interactive and adaptive gamified elements to students with unique needs.
Flores et al. (\citeyear{flores2019proposal}) presented a personalized model that assesses students’ skills using pretest, leveraging case-based reasoning and RL (Q-Learning) to optimize the sequence of learning resources, aiming to prevent issues like anxiety or boredom according to flow theory. Their work provides us a metric for determining success.

Kardan and Speily (\citeyear{kardan2010smart}) introduced an evolving web-based learning system capable of adapting itself to individual learners, by retrieving relevant content from the web, personalizing it based on learners’ characteristics and preferences, addressing challenges unique to lifelong learning scenarios through a hybrid machine learning technique. Lopes and Lopes (\citeyear{lopes2022review}) reviewed dynamic difficulty adjustment methods, another form of adaptation to learners.

The literature indicates a growing interest in adaptive learning systems; nonetheless, most studies implement adaptations offline, analyzing learner data retrospectively rather than adjusting content in real-time \cite{kabudi2021ai}. Previous work has also concentrated on the technical design of such systems, offering few rigorous comparisons between adaptive and non-adaptive approaches under equivalent instructional conditions. In addition, time-on-task (how long it takes a user to complete a task) remains the most prevalent metric for driving adaptivity, yet its standalone effectiveness has rarely been examined. This study addresses these gaps by evaluating a real-time online adaptive system against a non-adaptive baseline and by isolating a time-based measurement to assess its utility for guiding adaptation.

\section{Methodology and Implementation}
This section provides a detailed overview of our APSG framework together with the underlying algorithms that support it. Our research does not aim to invent a novel adaptive algorithm. However, our implementation incorporates several tailored modifications that may assist researchers wishing to replicate or extend the approach.

\subsection{APSG as a Puzzle Game}
We designed an APSG using the Unity engine to foster problem-solving by presenting players with pathfinding-based puzzles to solve. As players advance, the system selects puzzles calibrated to their current ability. The goal of each puzzle is to find the correct path from the start node to the end node while picking up various cargo pieces along the way. These puzzles were based on the puzzle game \textit{Cosmic Express} \cite{cosmicexpress}. We chose this game because of its simple, easy-to-learn rules and straightforward difficulty evaluation, which is essential in our adaptive system.

\begin{figure}[t]
\centering
\includegraphics[width=0.75\columnwidth]{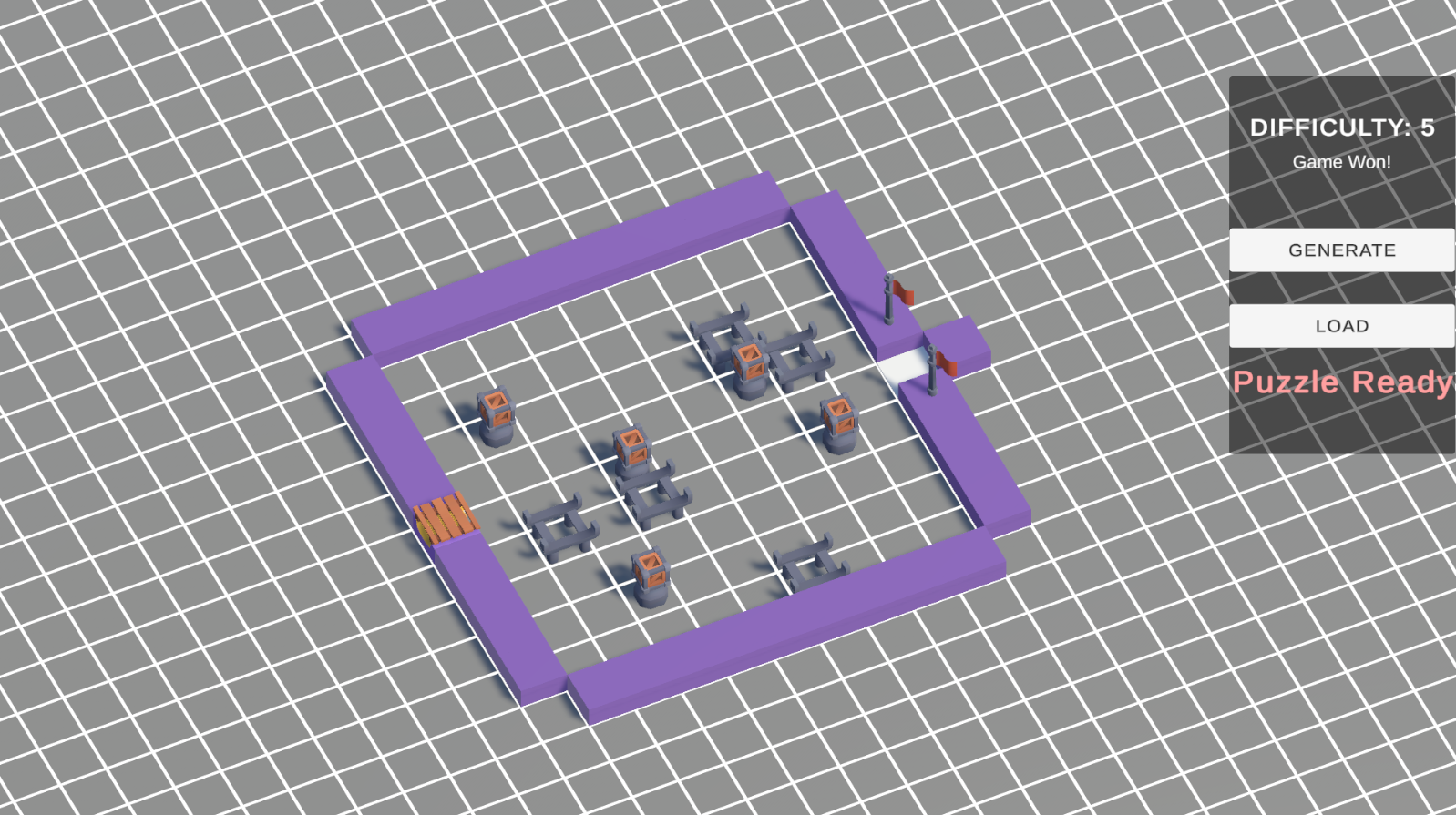} 
\caption{Example of a difficulty-5 puzzle.}
\label{fig1}
\end{figure}

\begin{figure}[t]
\centering
\includegraphics[width=0.75\columnwidth]{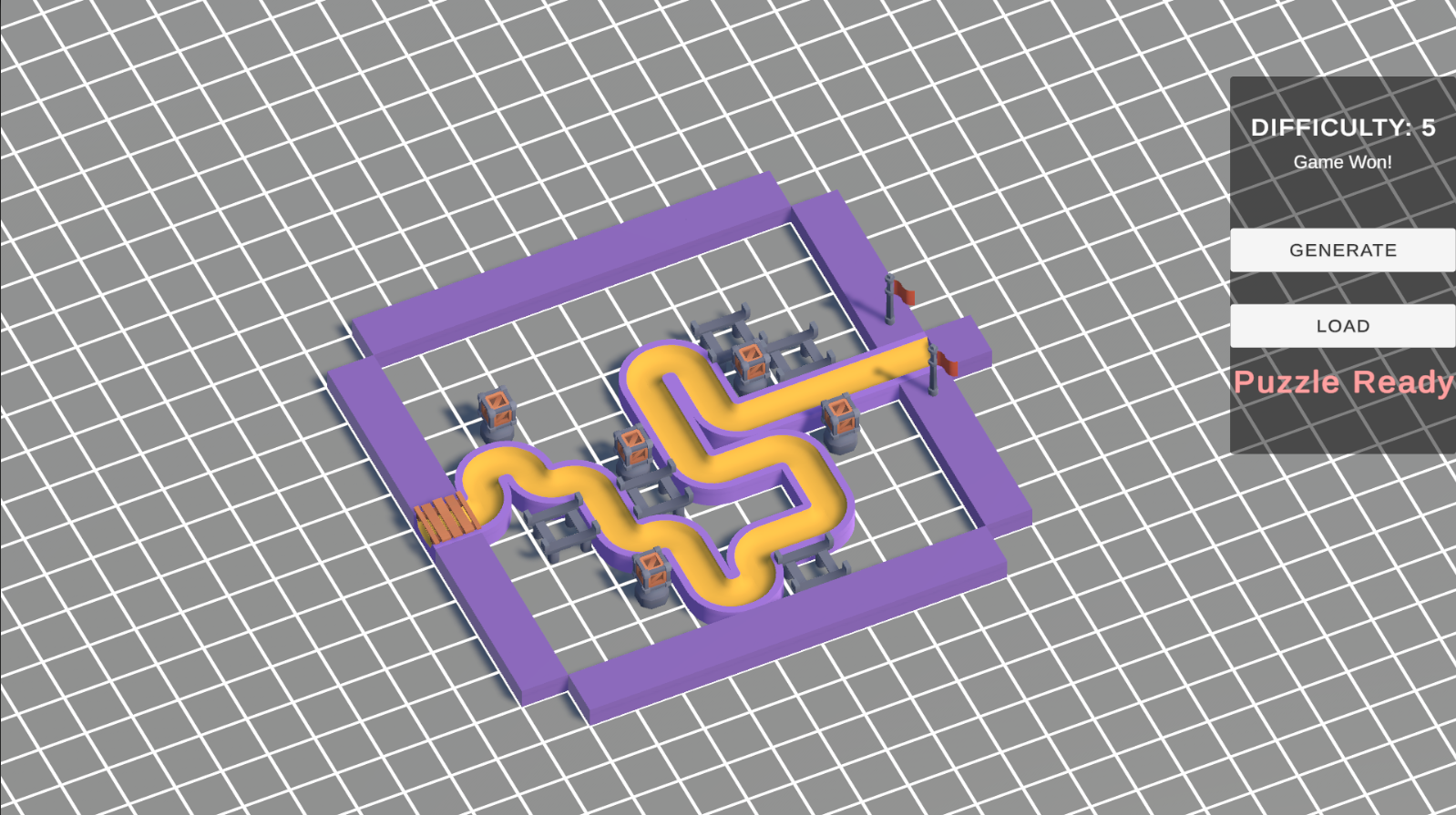} 
\caption{Example of a difficulty-5 puzzle with a solution path drawn by the player.}
\label{fig2}
\end{figure}

\begin{figure}[t]
\centering
\includegraphics[width=0.75\columnwidth]{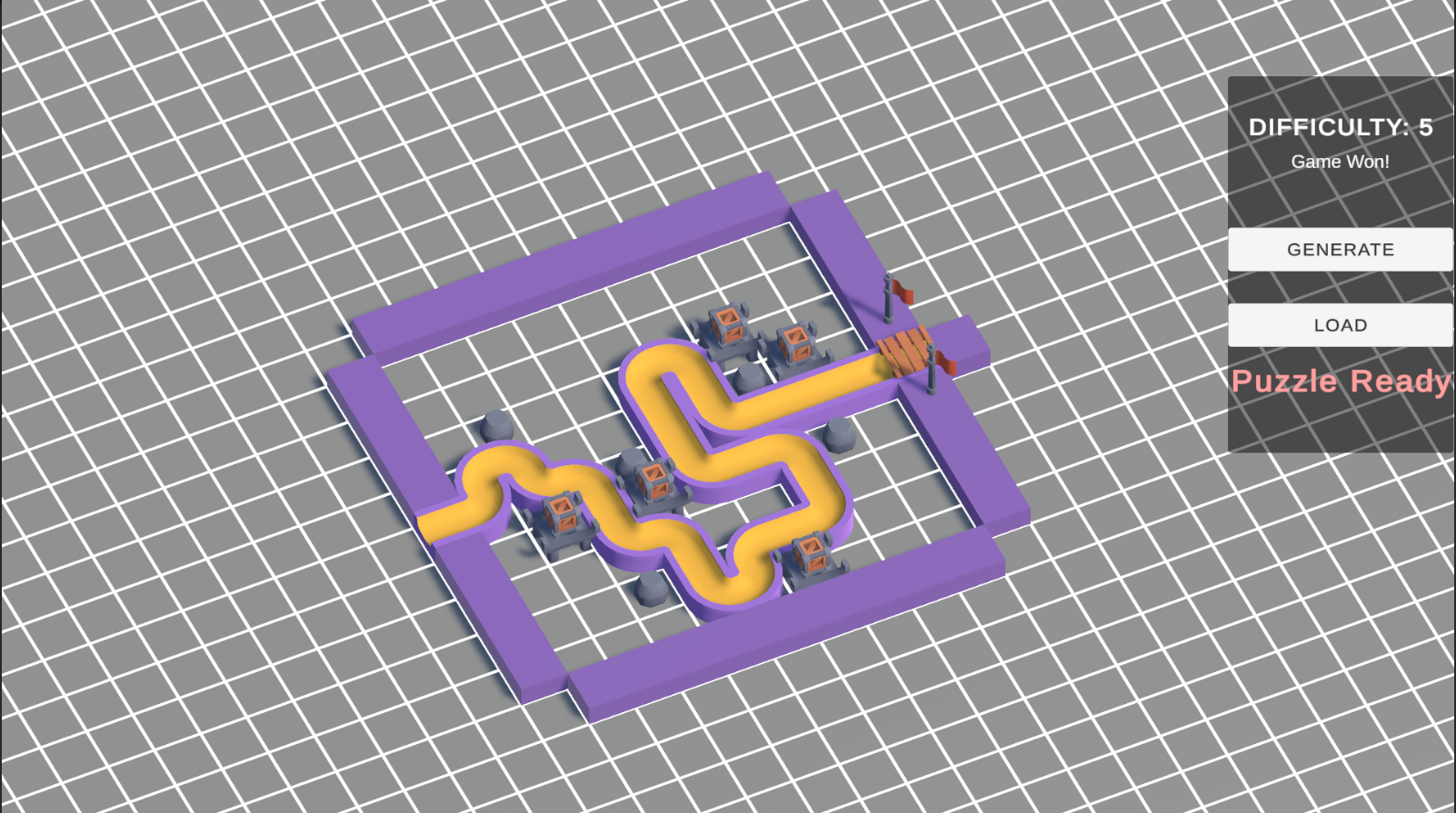} 
\caption{Example of a solved difficulty-5 puzzle, where all the cargo boxes are dropped off in the correct destinations.}
\label{fig3}
\end{figure}

The puzzle is represented on an $n\!\times\! n$ grid, in which the player needs to draw a path from the starting node to the end node (Figures \ref{fig1} and \ref{fig2}). Path pieces cannot overlap each other, the border of the puzzle, or ``special" points. There can only be a single path with no branches. The ``special" pieces (cargo) must be picked up at specific ``pickup" locations, and deposited at specific "dropoff" locations. Once a path is drawn, a ``container" automatically traverses the path one tile at a time, in the direction and order that the path has been drawn, and can carry exactly one cargo at any given time. Cargo is considered picked up or dropped off if at any time, this container passes a pickup or dropoff point adjacently. Importantly, the container can only ``hold" one cargo piece at a time, so careful planning of the path to ensure the order of operations for visiting each of the pickup or dropoff points needs to be considered. A particular puzzle is solved if:
\begin{itemize}
    \item The container can start at the start point and end at the end point, connected via a contiguous path.
    \item Once the container reaches the end point, there are no outstanding cargo pieces left at any of the pickup points.
    \item Each dropoff point contains exactly one cargo piece.
\end{itemize}

As the player is allowed to draw only one path at a time with no branches allowed, the container has only one path to follow. It either completes the puzzle or not, which determines whether the puzzle is solved. Figure \ref{fig3} shows the end result of the container going through the entire path, having picked up and dropped off cargo boxes along the way.

Puzzles can have many solutions (paths), often representing easier puzzles, or very few solutions, conversely representing difficult puzzles. The difficulty of the puzzle can be represented in a few different ways. First, the size of the grid. Larger grid sizes demand longer paths, which often add to the complexity of puzzles, particularly when combined with other difficulty metrics. Secondly, the number of pickup and dropoff locations. As these ``special" points increase in number, a given solution becomes more difficult to achieve as the possible order of visiting various locations increases in complexity. Finally, the specific location of special pieces. Special pieces might be placed in such a way that once visiting one, another becomes inaccessible, necessitating a redesign of the solution. Each puzzle that is presented to the user has been dynamically generated in real time, facilitated through the use of a genetic algorithm.

\subsection{Genetic Algorithm and Puzzle Generation}
The entirety of the puzzle generation pipeline is facilitated through the use of a genetic algorithm. Functional design decisions were made such that the GA can generate puzzles of varying difficulty, connect to the underlying player modeling methods, and provide generation speed to support ``real-time'' generation. The genetic algorithm generates a set of path and special points that represent the solution of the puzzle. These set of points are stored as an $n\!\times\! n$ character grid, based on the size of the puzzle. 
The GA (Algorithm \ref{alg:genetic_algorithm}) is designed in such a way to optimize difficulty to a given input difficulty ranging from one to ten. The GA is based on the NSFI-2POP structure \cite{scirea2020adaptive}.  This section describes the details of the system, where domain-specific choices had to be made that deviate from more traditional structures.
\begin{algorithm}[t]
\caption{Genetic Algorithm}
\label{alg:genetic_algorithm}
\textbf{Input}: Population $P$, size $N$, generation limit $G$\\
\textbf{Parameter}: Mutation rate $m$\\
\textbf{Output}: Best solution $B$

\begin{algorithmic}[1]
\STATE Initialize $gen \gets 0$, $B \gets \text{null}$, $bestFit \gets 0$
\WHILE{$gen < G$}
    \STATE $F \gets 0$, $maxFit \gets 0$, $best \gets \text{null}$
    \FOR{each $c$ in $P$}
        \STATE $f \gets \text{Fitness}(c)$
        \STATE $F \gets F + f$
        \IF{$f > maxFit$}
            \STATE $maxFit \gets f$
            \STATE $best \gets c$
        \ENDIF
    \ENDFOR
    \IF{$maxFit > bestFit$}
        \STATE $bestFit \gets maxFit$
        \STATE $B \gets \text{Clone}(best)$
    \ENDIF
    \STATE $gen \gets gen + 1$
    \STATE $P' \gets \{\}$
    \WHILE{$|P'| < N$}
        \STATE $p_1 \gets \text{Select}(P)$
        \STATE $p_2 \gets \text{Select}(P)$
        \STATE $(c_1, c_2) \gets \text{Crossover}(p_1, p_2)$
        \STATE $\text{Mutate}(c_1, m)$
        \STATE $\text{Mutate}(c_2, m)$
        \STATE Add $c_1, c_2$ to $P'$
    \ENDWHILE
    \STATE $P \gets P'$
\ENDWHILE
\STATE \textbf{Output:} $B$
\end{algorithmic}
\end{algorithm}

\subsubsection{Data Representation}
Puzzles are internally represented as a two-dimensional character grid. Encoded in this grid are various characters that represent specific elements of the puzzle (Table \ref{tab:character_table}). This representation allows for easy display of relevant puzzles and allows puzzles to be stored or loaded as needed.

\begin{table}[H]
    \centering
    \begin{tabular}{|c|c|}
    \hline
    \textbf{Character} & \textbf{Mapping}  \\
    \hline
         \# & empty space  \\
         X & path \\
         P & pickups \\
         D & dropoffs \\
         O & obstacles / border \\
         S & start point \\
         E & end point \\
        \hline
    \end{tabular}
    \caption{Puzzle Character Mapping.}
    \label{tab:character_table}
\end{table}

\subsubsection{Crossover Function}
Traditional genetic crossover functions take in two one-dimensional coded data points and then split them based on some criteria, to produce two corresponding child data points. The goal of the crossover function for the described APSG is to mix the path and grid configurations of two parent puzzles to create new children. Each child inherits part of the path and grid structure from one parent, and the rest from the other - aiming to blend traits and explore new puzzle variations. A random column is chosen as the crossover point, constrained to ensure that it is not too close to the puzzle edges while allowing parents to be merged non-symmetrically to increase variability. The child grids are built by the following two steps, forming a simple two-part combination:

\begin{enumerate}
    \item Copying the left side of the grid from one parent.
    \item Copying the right side of the grid from the other parent.    
\end{enumerate}
The path (solution) is split at the crossover column, where:
\begin{enumerate}
    \item Child 1 takes the first half of Parent 1's path and the second half of Parent 2's path.
    \item Child 2 does the reverse - taking the first half from Parent 2 and second half from Parent 1.
\end{enumerate}

\begin{algorithm}[t]
\caption{Crossover Function}
\label{alg:crossover_function}
\textbf{Input}: Parent puzzles $P_1$, $P_2$\\
\textbf{Output}: Child puzzles $C_1$, $C_2$
\begin{algorithmic}[1]

\STATE Select a crossover point
\STATE Swap puzzle sections between $P_1$ and $P_2$ to create $C_1$ and $C_2$
\STATE Adjust path using BFS
\STATE Adjust special points using distance-based metrics
\STATE Validate and finalize child puzzles

\end{algorithmic}
\end{algorithm}

Combining the paths often ``breaks" the solution, in either the path structure or special point representation. For the path, the crossover often directly creates gaps or overlaps, where two paths are no longer connected. As such, diagonal moves are corrected and a Breadth-First Search (BFS) is used to fill in missing steps between the broken path segments. For the ``special'' points, the children's pickup and dropoff points are recalculated using a distance-based metric that subdivides the path into segments, selecting candidate tiles within these segments. Pickups and dropoffs are then alternately placed at random among valid candidates, ensuring an equal number of each while avoiding forbidden positions. Paths are finalized by removing any duplicates or invalid moves and are checked for solvability. Algorithm \ref{alg:crossover_function} shows the entire process.

\subsubsection{Fitness Function}
The fitness function is used to evaluate the current analytical ``difficulty" of a given puzzle. It is essential that the genetic algorithm can optimize to any given input difficulty, including ``medium" or ``subjectively-defined" level difficulties. Difficulty was assigned a discrete 1-10 scale to align with our user study design, although the GA could optimize to scores on a broader range of difficulty levels. We define minimum and maximum values for various metrics that are then integrated to provide a current puzzle difficulty.
\begin{table}[t]
\centering
\begin{tabular}{|l|c|c|}
\hline
\textbf{Factor} & \textbf{Min Value} & \textbf{Max Value} \\
\hline
Path Length & 8 & 50  \\
Corners & 0 & 20  \\
Empty Space & 20 & 5  \\
Pickups & 1 & 12  \\
Orthogonal Pickups & 0 & 2 \\
\hline
\end{tabular}
\caption{Fitness function components and their value ranges. Target values are interpolated based on puzzle difficulty.}
\label{tab:fitness_table}
\end{table}

The total score is then calculated as a weighted sum of the fitness factors:
\[
score = \sum_{f \in F} \max(0, tar_f - |tar_f - act_f|) \times weight_f
\]

In this formula, \( F \) is the set of fitness components, \( tar_f \) is the target value for each factor, \( act_f \) is the actual observed value in the puzzle, and \( weight_f \) is the weight assigned to each component. Thus, we are able to obtain puzzles with a variety of ``scores", which can easily be mapped to a corresponding difficulty level between one and ten. This system can be easily tweaked to provide an extensive array of varying difficulties and their corresponding puzzles. However, the one to ten scale was implemented to facilitate the ease of use for a study. Table \ref{tab:fitness_table} shows the range of values.

\subsection{Adaptive Difficulty and Player Modeling}
The target optimization difficulty needed by the genetic algorithm is provided by a player modeling system. This system records information about the current state of the puzzles as well as the current metrics of the player. It then makes a suggestion in terms of difficulty adjustment. The system suggests the difficulty of the puzzles should:
\begin{enumerate}
    \item Increase, as the puzzles are too easy, or,
    \item Decrease, as the puzzles are too difficult, or,
    \item Neither increase or decrease, as the puzzles are a suitable difficulty. 
\end{enumerate}A mixture of hard constraints (constraints that must be validated for a difficulty transition) and soft constraints (constraints that help determine finer aspects of the transition) are integrated into the player model. These constraints are based on the following measured metrics of the player and puzzle state:
\begin{enumerate}
    \item Time taken to reach solution
    \item Number of attempts before solution
    \item Number of backtracks (removing a portion of the puzzle to try again)    
    \item Number of times the puzzle state was reset
    \item Number of times the puzzle was almost solved (missing less than 25\% of special points)
\end{enumerate}
Currently, the only metric with relation to hard constraints is the \textit{Number of attempts before a solution}, as we are setting a hard limit on the number of attempts the player can have before they are unable to increase the difficulty. This hard constraint enforces minimum playability requirements, ensuring the model does not increase the difficulty when a large number of attempts have been made. The rest of the metrics feed in to the soft constraint calculation, to inform the overall player model output, allowing the model to adjust difficulty smoothly.
\begin{algorithm}
\caption{Calculate Soft Constraint Score}
\label{alg:soft_constraints}
\textbf{Input}: Player metrics: backtracks $B$, near-solves $N$, resets $R$, time taken $T$ \\
\textbf{Output}: Soft constraint score $S_s$
\begin{algorithmic}[1]

\STATE $B \gets B - 1$ \COMMENT{Ignore initial start count}

\STATE $S_s \gets 0$

\IF{$B < B_{\text{threshold}}$}
    \STATE $S_s \gets S_s + |10 - B| \times W_B$
\ELSE
    \STATE $S_s \gets S_s - B \times W_B$
\ENDIF

\IF{$N < N_{\text{threshold}}$}
    \STATE $S_s \gets S_s + |5 - N| \times W_N$
\ELSE
    \STATE $S_s \gets S_s - N \times W_N$
\ENDIF

\IF{$R < R_{\text{threshold}}$}
    \STATE $S_s \gets S_s + |5 - R| \times W_R$
\ELSE
    \STATE $S_s \gets S_s - R \times W_R$
\ENDIF

\STATE $S_s \gets S_s - T \times W_T$

\RETURN $S_s$

\end{algorithmic}
\end{algorithm}

The player model takes in the soft constraints score in tandem with the validity of passing hard constraints, and suggests a new difficulty of puzzle.  Algorithm \ref{alg:soft_constraints} shows how the soft constraints are used to calculate a score.

\subsection{Puzzle Evolution and Metrics}

\begin{figure}[t]
\centering
\includegraphics[width=0.75\columnwidth]{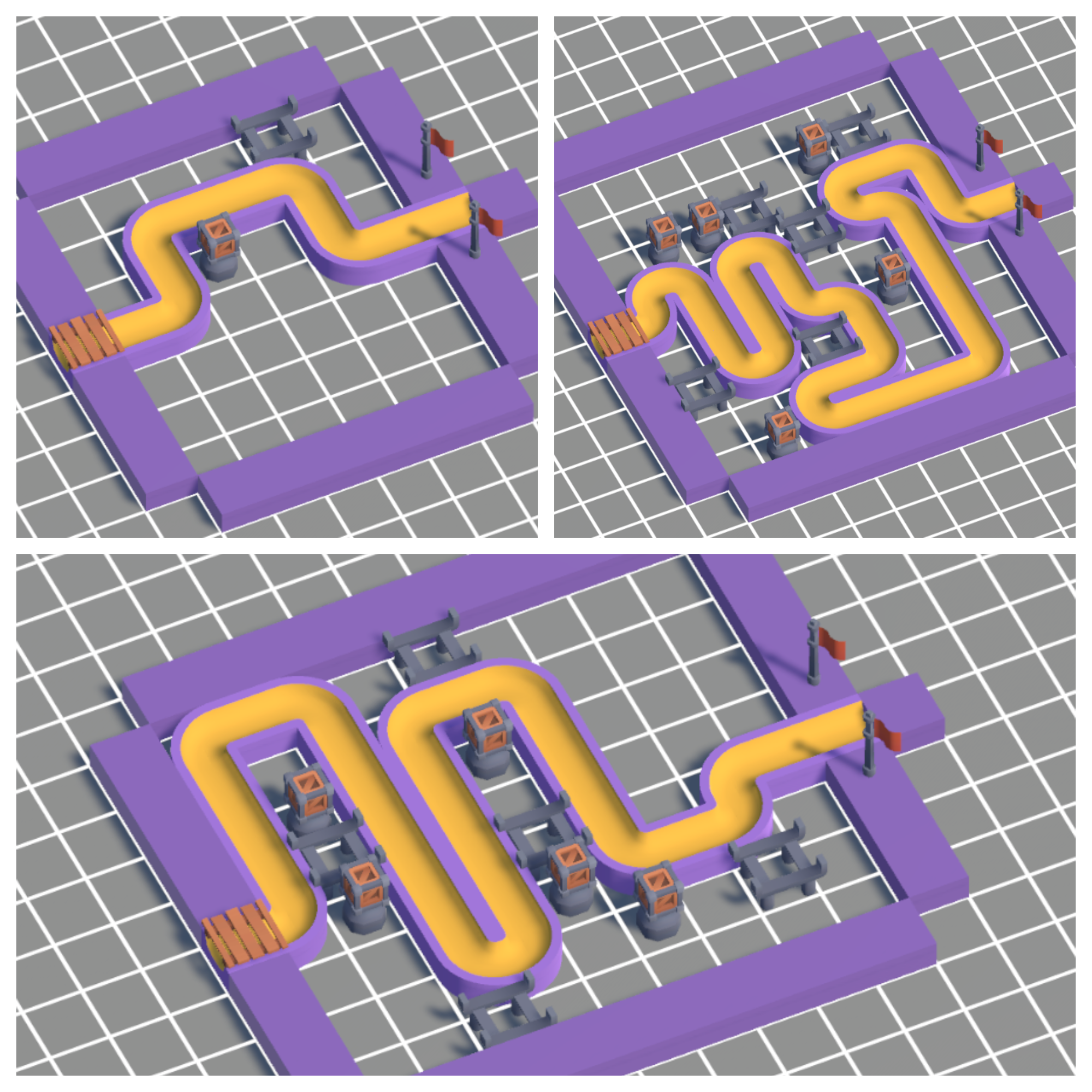} 
\caption{Difficulties 1 (top left), 5 (bottom) and 10 (top right) puzzles.}
\label{fig4}
\end{figure}

\begin{figure}[t]
\centering
\includegraphics[width=0.75\columnwidth]{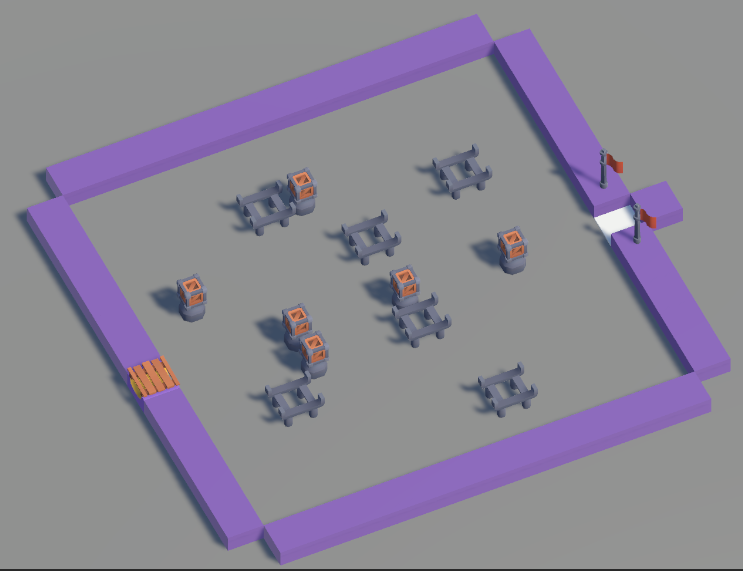} 
\caption{Example of large grid puzzle.}
\label{fig5}
\end{figure}

\begin{figure}[t]
\centering
\includegraphics[width=0.75\columnwidth]{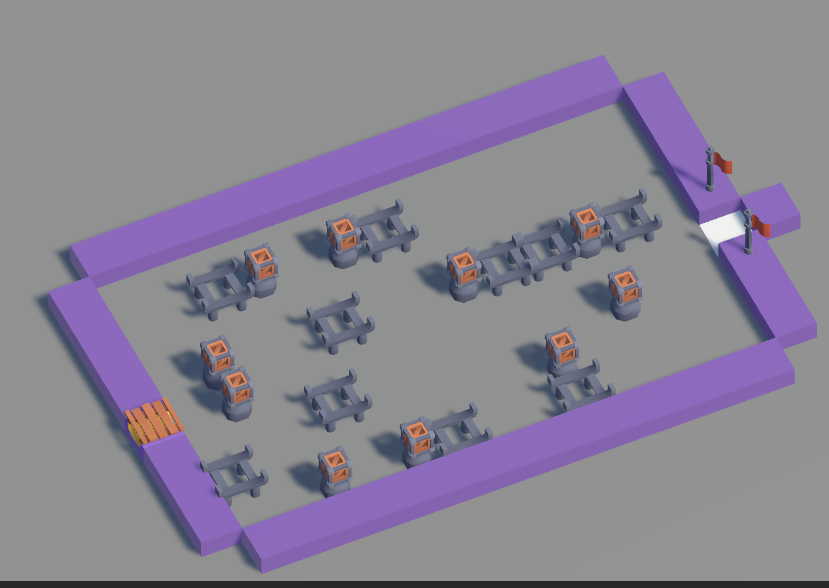} 
\caption{Example of narrow puzzle.}
\label{fig6}
\end{figure}

Figure \ref{fig4} showcases 3 generated puzzles of increasing difficulty, showcasing that the system is able to produce varying puzzles configurations, across varying grid sizes. As the difficulty increases, we see transitions from linear, simple to understand and ``empty" solutions, to those with more complex and sprawling paths, that often incorporate more of the available grid space. Deeper thought is required in terms of ``harder" puzzles, as there are many avenues that will not work, and often require more problem solving, critical thinking, and trial and error. 
Further, the genetic system is able to generate puzzles with unique dimensions, and varying complexity. Figure \ref{fig5} showcases a puzzle with large amounts of empty space between various points, allowing for multiple solutions. On the other hand, Figure \ref{fig6} showcases a complex puzzle along a narrow grid, in which there exists less opportunity for varying solutions. 
The various genetic algorithm parameters are easily controlled to produce a variety of unique puzzles with varying complexity. To provide a decent tradeoff between runtime complexity and varying puzzles, we have tested various parameters. Table \ref{tab:params} shows parameters selected for the user study, selected to ensure puzzles are generated with reasonable runtimes while providing sufficient variability. 
\begin{table}[H]
    \centering
    \begin{tabular}{|c|c|}
    \hline
    \textbf{Parameter} & \textbf{Value}  \\
    \hline
         Population Size & 300 \\
         Crossover Rate & 80\%  \\
         Generation Limit & 10 \\
         Maximum Grid Size & 10x10 \\
        \hline
    \end{tabular}
    \caption{Parameters Used for User Study.}
    \label{tab:params}
\end{table}
Utilizing the above parameters, we are able to generate a puzzle of any difficulty in approximately 7 seconds of runtime, with more difficult and complex puzzles taking longer than trivial ones. This allows for online puzzle generation, while maintaining relatively low load times. As the population size, generation limit and maximum grid size are increased, extremely complex and large puzzles are able to be generated, however, runtime is affected substantially.

\section{User Study Results and Discussion}

\subsection{User Study}

We have received ethics approval from the Research Ethics Board at our institution to conduct a user study. Participation was completely voluntary. Participants were compensated with a \$20 Amazon gift card and recruited from a computer science graduate student mailing list. The goal of this exploratory study is to assess the functionality of the system, and to determine if the combination of PCG and player modeling provided an enriching problem solving environment. This exploratory study provides us initial analytical data with regards to the ``feeling" of the system, and the presentation of particular puzzles at particular times. This experiment consisted of participants playing through ten puzzles in each of three different versions of the APSG. 
\begin{enumerate}
    \item \textbf{Standard}, in which our most complex player modeling implementation was provided.
    \item \textbf{Increasing}, a variation in which the puzzle always is increasing by one difficulty level, regardless of player performance.
    \item \textbf{Time-based}, in which the only player modeling metric fed into the modeling system was the time taken to solve a particular puzzle (the first metric from the list before).
\end{enumerate}

Each participant played through all three versions. The order of the version presentation was switched between participants, to ensure that the aggregate results were not skewed based on users learning the puzzles and performing better on later versions. 

Following the gameplay session, users were asked to provide feedback via a short questionnaire. Basic demographic such as age, gender and university major were recorded. Detailed questions regarding the gameplay experience, problem-solving learning experience and usability metrics were provided to the users. Finally, we recorded various metrics regarding the individual dynamic difficulty modes, as to determine the effectiveness of various metrics in the player modeling system. All questions were provided as Likert scales, either on a number system for determining player modeling metrics, or agree/disagree scales for the subjective usability questions. 

\begin{figure}[t]
\centering
\includegraphics[width=0.75\columnwidth]{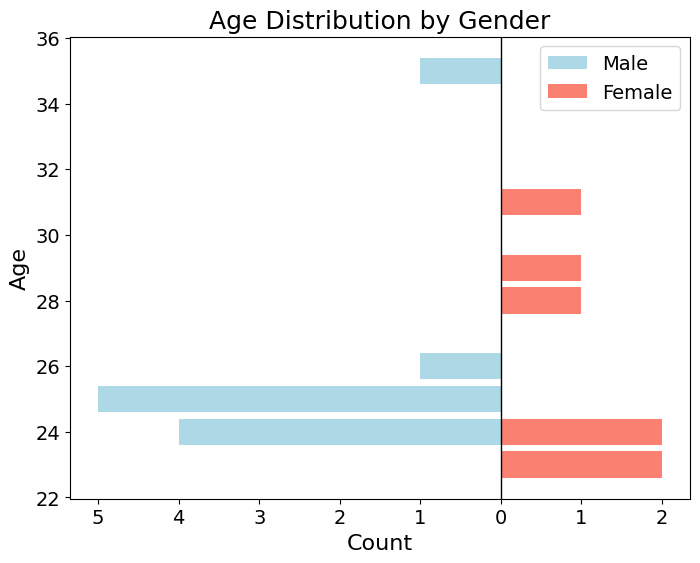} 
\caption{Age and Self-Reported Gender Distribution.}
\label{fig7}
\end{figure}

\subsection{Results and Discussions}
A total of 18 participants took part in the study, representing a range of ages and self-reported genders (Figure \ref{fig7}). Overall, responses to the APSG as a whole were positive. The vast majority of participants found the experience to be intellectually engaging, with \textbf{89\%} either \textit{agreeing} or \textit{strongly agreeing} that the game was stimulating. Additionally, \textbf{94.5\%} of participants reported that the game required the use of \textit{problem-solving skills}, reinforcing the notion that the gameplay commanded thoughtful engagement. A slightly smaller, though still significant, proportion - \textbf{72.2\%}, felt the game required \textit{critical thinking skills}. Taken together, these responses suggest that the APSG successfully delivered a cognitively demanding experience, aligning more with the goals of serious games rather than purely entertainment-focused gameplay.

To answer our research questions, we asked each participant to rate the following statements on a scale of 1 to 10, for each version of the game:
\begin{enumerate}
    \item The generation system reduced frustration with puzzle solving.
    \item The generated puzzles were of the right difficulty.
    \item There was a noticeable change in puzzle difficulty.
    \item I felt a sense of progression (earlier puzzles were easier than later puzzles).
\end{enumerate}

\begin{figure}[t]
\centering
\includegraphics[width=0.75\columnwidth]{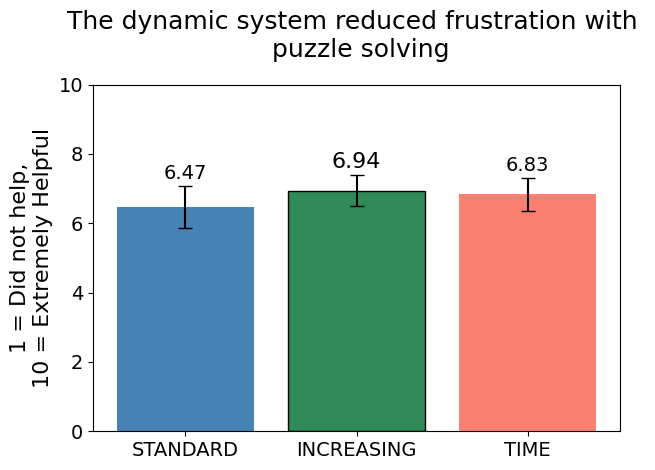} 
\caption{Reduced frustration.}
\label{fig11}
\end{figure}

\begin{figure}[t]
\centering
\includegraphics[width=0.75\columnwidth]{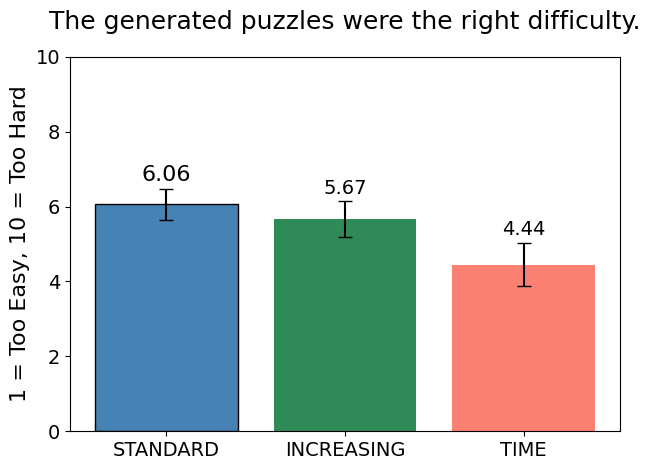} 
\caption{Suitable Difficulty.}
\label{fig12}
\end{figure}

\begin{figure}[t]
\centering
\includegraphics[width=0.75\columnwidth]{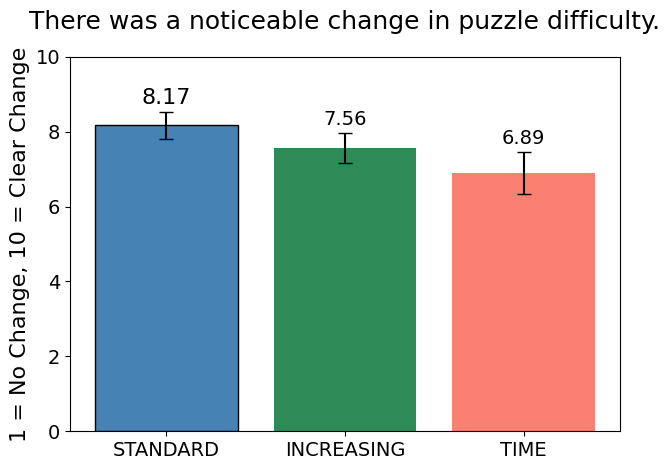} 
\caption{Noticeable Change in Puzzle Difficulty.}
\label{fig13}
\end{figure}

\begin{figure}[t]
\centering
\includegraphics[width=0.75\columnwidth, height=5.05cm]{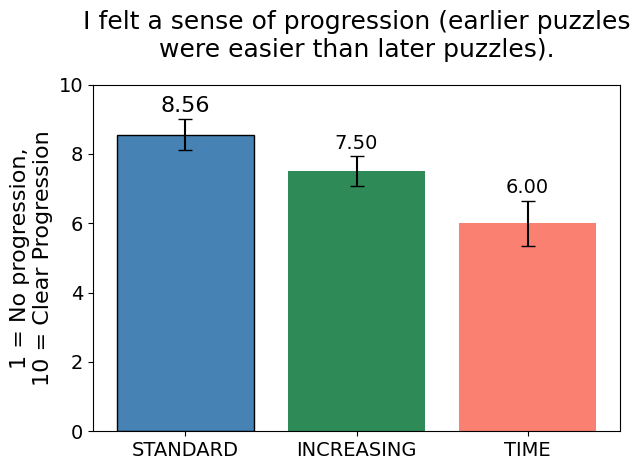} 
\caption{Progression of the Game.}
\label{fig14}
\end{figure}

\begin{figure*}[t]
\centering
\includegraphics[width=1.4\columnwidth, height=5.05cm]{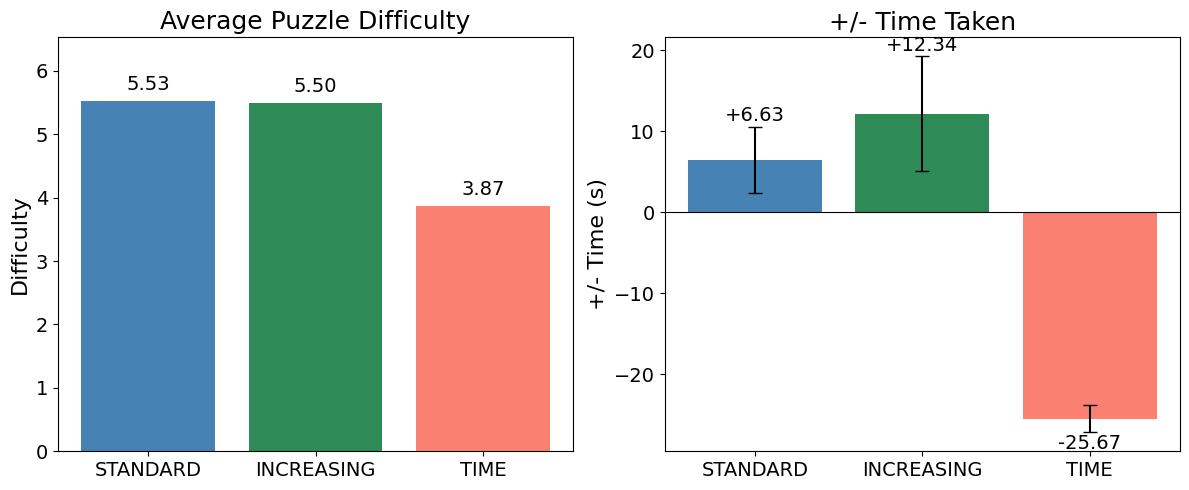} 
\caption{Average Puzzle Difficulty and Average Deviation in Completion Time Per Puzzle compared to the benchmark.}
\label{fig15}
\end{figure*}

\subsubsection{Frustration (RQ1)}

We use the result of Statement 1 to answer RQ1. Across all three versions of the system, participant responses revealed a consistent moderate reduction of frustration while solving puzzles (see Figure \ref{fig11}). Average scores tended towards the higher end of the scale, indicating lower frustration, with the Increasing and Time-based versions performing slightly better than the Standard (Standard = 6.47, Increasing = 6.94, Time-based = 6.83). However, there were no statistically significant differences between group means as determined by one-way ANOVA: F(2, 51) = 0.072, p = 0.93. These results suggest that all three approaches had a similar positive impact on player experience by maintaining a manageable challenge level, and players did not feel that the adaptive difficulty system had an impact. However, the absence of a static baseline comparison limits our ability to quantify the system's impact. Future studies could include a static version of the game as a baseline to measure frustration levels more strictly.

\subsubsection{Difficulty and Progression (RQ2)}

To answer RQ2, we examine the results from Statements 2 to 4. When examining perceived or suitable difficulty, participants were asked to rate the puzzles on a scale where one extreme represented puzzles that were too easy, and the other extreme represented puzzles that were too difficulty (see Figure \ref{fig12}). We hypothesized that the optimal difficulty would fall just above the midpoint of the scale, signaling that players found the puzzles to be challenging, but not overwhelming. This hypothesis was generally supported by the results, with the Standard and Increasing versions hovering near the expected sweet spot (Standard = 6.06, Increasing = 5.67), seemingly indicating that these models struck a decent balance in challenge. Conversely, the Time-based model received a noticeably lower score (Time-based = 4.44), suggesting that the puzzles may have skewed towards being too easy. ANOVA shows a difference between the means: F(2, 51) = 3.447, p = 0.039.  A post-hoc t-test with Bonferroni correction between Standard and Time-based models shows significant differences and large effect size (p = 0.004, Cohen's d = 0.869), while there are no significant differences between Standard and Increasing models (p = 0.075, Cohen's d = 0.330).

To assess whether the difficulty changed in a noticeable or meaningful way across the course of the session, we analyzed participant responses to the statement targeting perceived variation in puzzle difficulty (see Figure \ref{fig13}). Ideally, a well-designed adaptive system should present a clear track of increasing challenge. The results indicated that the Standard model most effectively conveyed this change (Standard = 8.17), followed closely by the Increasing model (Increasing = 7.56). The Time-based model again trailed behind (Time-based = 6.89). However, ANOVA shows no statistical significance: F(2, 51) = 1.979, p = 0.149.

Finally, participants were asked whether they felt a sense of progression throughout the sequence of puzzles - a key factor in our hypothesis of player engagement (see Figure \ref{fig14}). Responses situated closely with our expectations: the Standard model once again led in perceived progression (Standard = 8.56), with the Increasing model following behind (Increasing = 7.5). The Time-based version scored the lowest (Time-based = 6.0), reinforcing patterns seen in previous measures. ANOVA shows a difference between the means: F(2, 51) = 5.758, p = 0.005.  A post-hoc t-test with Bonferroni correction between Standard and Time-based models shows significant differences and large effect size (p $<$ 0.001, Cohen's d = 1.083). While there are no significant differences between Standard and Increasing models (p = 0.074), there is a moderate effect size (Cohen's d = 0.567). These findings suggest that while all systems incorporated some degree of progression, the Standard and Increasing models provided a more coherent sense of advancement. 

\subsubsection{Gameplay Data}

In addition to subjective, questionnaire based feedback, analytical gameplay data was collected in the form of detailed user logs for each participant across versions. These logs recorded the specific puzzles completed, the time taken to solve each one, their difficulty, and key player modeling metrics used during adaptive generation. To further assess the effectiveness of each version, we analyzed two core metrics: the \textbf{average puzzle difficulty} presented to players, and their \textbf{average deviation in completion time}, relative to a unified benchmark. This benchmark was obtained by averaging the completion times across all puzzles in the study, creating a reference point where each version's performance could be compared. For each version, we then calculated the average time offset, which is either above or below this benchmark, offering insight into how long participants took to complete puzzles relative to the overall average. Figure \ref{fig15} visualizes these comparisons, showcasing both average puzzle difficulty and average time deviation per version. Here, positive values indicate participants took longer than the benchmark, while negative values reflect faster completions, again relative to the benchmark.

The results indicate several potential takeaways. Both the Standard and Increasing models generated puzzles with similar average difficulty levels (Standard = 5.53, Increasing = 5.50). However, participants completed puzzles in the Standard version much faster, with an average time deviation of \textbf{+6.63 seconds}, compared to \textbf{+12.34 seconds} in the Increasing model. This suggests that although both systems presented similarly difficult puzzles, the Standard version enabled players to solve them more efficiently, potential reflecting more suitable pacing, or better alignment with player ability over time. The Time-based version conversely showed a significantly different pattern. While its average time deviation was much lower, at \textbf{-25.67 seconds}, this apparent speed came at the cost of overall puzzle challenge: the average puzzle difficulty for this version was notably lower (Time-based = 3.87). In other words, this indicates that players in the Time-based model were solving puzzles much faster than the benchmark - but likely due to the system failing to escalate challenges effectively. As a result, participants rarely encountered higher-difficulty puzzles, that might have required longer engagement.

\subsubsection{Time as a Metric (RQ3)}

When considering the subjective, questionnaire-based measures in tandem with our methodological log-based analysis, a pattern seemingly begins to emerge. Across multiple criteria - noticeable change in puzzle difficulty (Figure \ref{fig13}), perceived difficulty suitability (Figure \ref{fig12}), and sense of overall progression (Figure \ref{fig14}), the Standard model consistently received the highest ratings. The Increasing model showed promise, but was slightly less effective in creating a smooth and timely experience, evident in the longer completion times. Meanwhile, the Time-based model consistently underperformed across both subjective and analytical dimensions, suggesting that time alone as a metric may not be well-suited for adaptive puzzle generation, particularly in its current form. 

Taken together, these initial findings suggest that the Standard player model, which incorporates complex and more nuanced tracking of user performance and engagement, potentially offers the most balanced and effective foundation for adaptive difficulty adjustments. However, we note that the Standard and Increasing models do not have statistical significant differences across various measures, indicating that further studies must be conducted. These insights suggest further development and refinement of the Standard model as a potential core approach for adaptive puzzle generations in future iterations of the APSG.

\section{Conclusions and Future Work}
This paper proposes a novel use of a genetic algorithm in tandem with an adaptive player modeling system to train students with problem solving. The presented APSG is capable of creating a wide variety of different puzzles, with varying difficulties and complexities. Further, it is able to dynamically adapt the difficulty of presented puzzles to adapt to the current user, and can present the player with a progression of simple puzzles to complex ones. Additionally, the pilot user study results suggest positive notions to the APSG as a whole, with the insight that having only time as a metric produces notably worse outcomes overall.

Through this research, we present a customized design of a genetic algorithm in an APSG in which puzzles of varying difficulties can be generated. The integration of user data with AI-driven technologies aims to foster strong engagement by allowing users to be fully immersed while remaining in control.

\subsection{Limitations}
While our results show promising indications of the impact of the model, further work is necessary. First, our results are limited by the relatively small number of participants. The only metric that we have singled out is time-on-task, based on its usage in prior literature. The selection for the time-on-task threshold was chosen at a group level despite an unknown participant population. A full ablation study could examine other metrics in the player model and their effectiveness, with future work exploring the individual impacts of each metric rather comparing versions.
Additionally, various validity concerns should be considered. Internal, external, and construct validity may be influenced by nuanced aspects such as player skill, design-driven parameter choices, self-reporting metrics, and the limited scope of puzzle types and participant population. Future work should consider and refine measurement approaches and the affects of underlying variables, while also considering the system's generalizability.

The player modeling system could also benefit from more advanced metrics, and the calculation thereof, particularly those concerning emotional states. If such a system could record the emotional states of users, through integrated reality technologies, for instance, eye-tracking, it could then be analyzed in such a way to provide an emotional metric with regards to difficulty, rather than what can be measured strictly from playing through the puzzles. Further, the player modeling system might benefit from various AI-techniques, instead of the simplistic approach we used, perhaps reinforcement-learning or machine-learning based approaches. This would allow for a more exact model of the player, and as such, could suggest more suitable difficulties. 

Our systematic approach to adaptive puzzle generation has broader impact beyond the current work. It could be implemented across more complex and non-linear puzzles. For instance, puzzles that deal with topics other than pathfinding such as mathematical or logic-based puzzles present a unique opportunity for expanding this system. Additionally, such a system has yet to be used in a non-game environment. For example, a similar system that instead optimizes practice problems for grade-school or similar curricula, particularly in the math or programming domains might warrant further exploration. Finally, the development of a similar system targeted towards diverse audiences would be beneficial to encompass all learners. Particularly, a well-designed system could be aimed towards those with learning disabilities, aiming to adapt to their specific and exacting educational needs.

\section{Acknowledgements}
This research was supported by the Natural Sciences and Engineering Research Council of Canada (NSERC) Discovery Grant, the NSERC Canada Graduate Scholarships (CGS-M), and the Alberta Graduate Excellence Scholarship (AGES). We thank members of the Serious Games Research Group and the anonymous reviewers for their feedback.

\bigskip

\bibliography{aaai25}

\appendix

\end{document}